\documentclass[10pt,twocolumn,letterpaper]{article}

\usepackage{cvpr}              % To produce the CAMERA-READY version
\usepackage{times}
\usepackage{epsfig}
\usepackage{graphicx}
\usepackage{amsmath}
\usepackage{amssymb}
\usepackage{bm,multirow,arydshln,subcaption,booktabs,makecell}
\definecolor{cvprblue}{rgb}{0.21,0.49,0.74}
\usepackage[pagebackref,breaklinks,colorlinks,allcolors=cvprblue]{hyperref}

\def\paperID{13202} % *** Enter the Paper ID here
\def\confName{CVPR}
\def\confYear{2025}

\title{Recurrent Feature Mining and Keypoint Mixup Padding \\ for Category-Agnostic Pose Estimation}

\author{Junjie Chen\textsuperscript{1}\quad Weilong Chen\textsuperscript{1}\quad Yifan Zuo\textsuperscript{1}\quad Yuming Fang\textsuperscript{1}\thanks{Corresponding author}\\
\textsuperscript{1} Jiangxi University of Finance and Economics\\
{\tt\small \{chenjunjie,chenweilong,zuoyifan,fangyuming\}@jxufe.edu.cn}
}

\begin{document}
\maketitle
\begin{abstract}
Category-agnostic pose estimation aims to locate keypoints on query images according to a few annotated support images for arbitrary novel classes.
Existing methods generally extract support features via heatmap pooling, and obtain interacted features from support and query via cross-attention.
Hence, these works neglect to mine fine-grained and structure-aware (FGSA) features from both support and query images, which are crucial for pixel-level keypoint localization.
To this end, we propose a novel yet concise framework, which recurrently mines FGSA features from both support and query images.
Specifically, we design a FGSA mining module based on deformable attention mechanism.
On the one hand, we mine fine-grained features by applying deformable attention head over multi-scale feature maps.
On the other hand, we mine structure-aware features by offsetting the reference points of keypoints to their linked keypoints.
By means of above module, we recurrently mine FGSA features from support and query images, and thus obtain better support features and query estimations.
In addition, we propose to use mixup keypoints to pad various classes to a unified keypoint number, which could provide richer supervision than the zero padding used in existing works.
We conduct extensive experiments and in-depth studies on large-scale MP-100 dataset, and outperform SOTA method dramatically (+3.2\%PCK@0.05).
The code of our \textbf{F}eature \textbf{M}ining and \textbf{M}ixup \textbf{P}adding method (FMMP) is avaiable
at \url{https://github.com/chenbys/FMMP}.
\end{abstract}
\section{Introduction} \label{sec:intro}
Pose estimation is a fundamental and significant computer vision task, which aims to produce the locations of pre-defined semantic part of object instance in 2D image.
Recently, it has received increasing attention in the computer vision community due to its wide applications in virtual reality, augmented reality, human-computer interaction, robot and automation.
However, most pose estimation methods are trained with category-specific data and thus cannot be applied for novel classes, especially when they have different keypoint classes.
Therefore, category-agnostic pose estimation (CAPE) \cite{POMNet} is introduced to locate target keypoints for arbitrary classes given a few support images annotated with keypoints.

\begin{figure}[t]
\begin{center}
\includegraphics[width=1\linewidth]{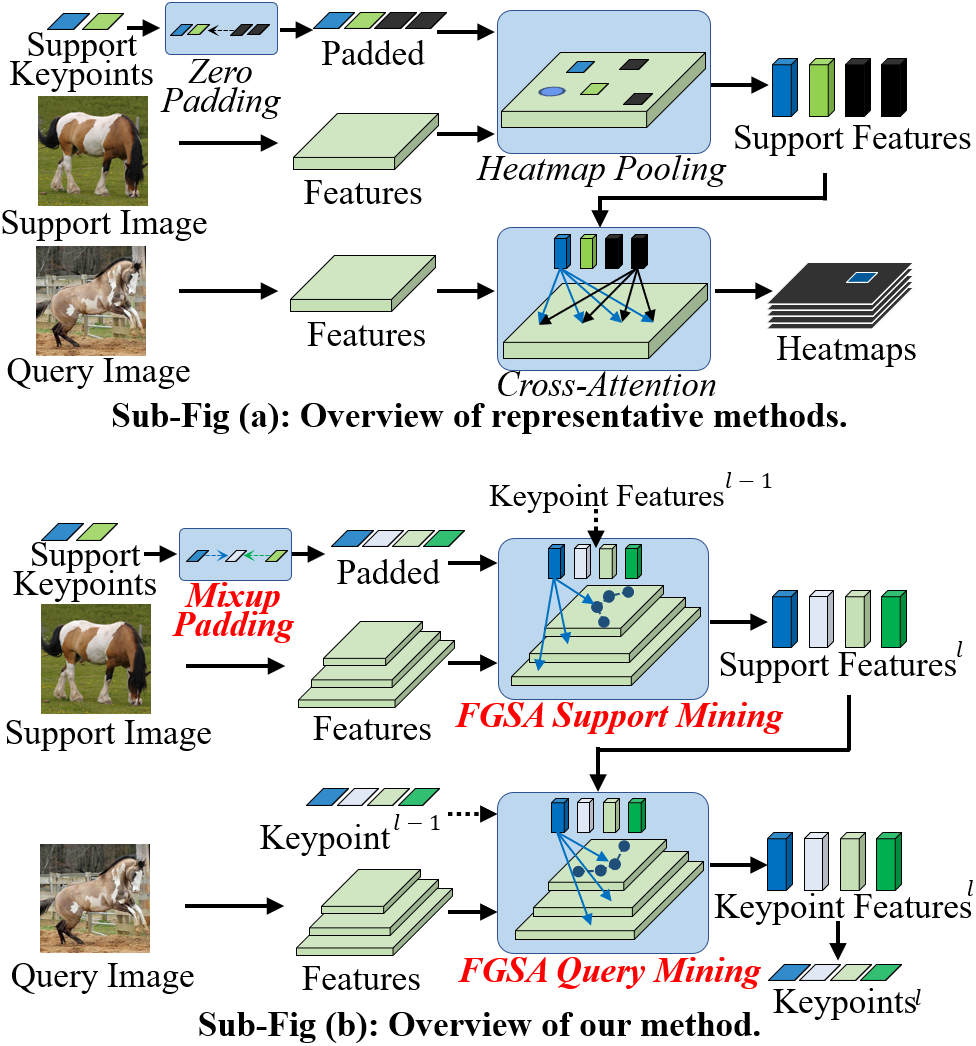}
\end{center}
\vspace{-17pt}
\caption{Overview of representative methods and ours.
(a): Representative methods use zero padding to align keypoint number, and rely on the heatmap pooling and cross-attention on single-scale features to produce support features and query estimations.
(b): Our method uses mixup padding, and recurrently mines FGSA features from multi-scale features of support and query images, which can produce better support features and query estimations.
}
\label{fig:overview}
\end{figure}

In CAPE, the query images come from novel classes, and thus the target keypoints to be estimated are determined by the annotated keypoints on support images.
Consequently, extracting high-quality features from both support images and query images are fundamental issues in CAPE.
As shown in Fig.~\ref{fig:overview} (a), representative methods \cite{POMNet,CapeFormer} firstly extract support features via feature map pooling weighted by a heatmap drawing annotated keypoint,
% (summarized as heatmap pooling or soft ROI pooling \cite{CapeFormer})
and then enable the interaction between support features and query feature map by cross-attention.
However, the pooling or cross-attention over single feature map are coarse-grained and inadequate for pixel-level keypoint localization.
Recent methods propose to complement support features with prototypical \cite{MetaPoint,ESCAPE}, global \cite{SCAPE} or structural \cite{Ru1,SDPNet} information, but how to extract fine-grained and structure-aware features from both support and query images remains unexplored.

In this paper, we propose a novel yet concise framework to recurrently mines fine-grained and structure-aware (FGSA) features from both support and query images.
Specifically, our framework consists of multiple stacked layers, and the pipeline of each layer is shown in Fig.~\ref{fig:overview} (b).
Generally, each layer firstly extracts FGSA features from the feature pyramid of support images according to the given support keypoints, and then absorbs FGSA features from query images to produce high-quality keypoint features.
In this way, our framework could recurrently refine the support features, keypoint features and estimated keypoints, and thus estimate more precise keypoints.

To mine FGSA features from both support and query images, we design a FGSA mining module based on deformable attention \cite{DeformDETR}, which is more flexible and effective against the heatmap pooling or cross-attention used in \cite{POMNet,CapeFormer,SCAPE}.
On the one hand, we set keypoints as reference points to mine fine-grained features from feature pyramid, which is especially beneficial for pixel-level keypoint localization.
On the other hand, we offset the attention heads of each keypoint to its linked keypoints, which could adaptively extract structure-aware features to facilitate the keypoint localization.
Therefore, our module could produce high-quality support features and keypoint features.

Besides, an inevitable issue in CAPE is keypoint padding, because all classes could have different numbers of keypoints.
Previous works intuitively align the keypoint number of various classes by zero padding, \emph{i.e.}, generating meaningless keypoints with zero weight/visibility.
In spired by Mixup \cite{mixup} and DensePose \cite{guler2018densepose}, we propose to align the keypoint number by mixup padding, \emph{i.e.}, generating dense keypoints by random mixing two linked keypoints.
Therefore, our mixup padding could provide richer supervision and enable our model to learn denser semantics of poses.

For the experimental setting, we follow previous works \cite{POMNet,CapeFormer,MetaPoint} to conduct experiments on MP-100 dataset \cite{POMNet}. 
The quantitative and qualitative experiments demonstrate the effectiveness of our proposed method.
Our contributions can be summarized as:
\textbf{1)} We propose a novel framework to recurrently mine fine-grained and structure-aware features from both support and query images, which could extract better support features and locate more precise keypoints on query images.
\textbf{2)} We propose a feature mining module based on deformable attention to integrally mine fine-grained and structure-aware features.
\textbf{3)} We propose a mixup padding strategy to provide richer supervision for model to learn denser semantics.
\textbf{4)} We conduct compresentive experiments on MP-100 dataset, and our model outperforms SOTA method dramatically (+3.2\% PCK@0.05). 

\section{Related Works}

\subsection{Category-Specific Pose Estimation}
Pose estimation is a fundamental and important vision task, aiming at detecting pre-defined keypoints of objects in image. 
Most existing methods are class-specific, \emph{i.e.}, focusing on estimating keypoints for single and specifc class, \emph{e.g.}, humans \cite{humanpose1,humanpose2}, animals \cite{animalpose1,animalpose2}, or vehicles \cite{carpose1,carpose2}. 
Technically, existing pose estimation approaches can be broadly divided into three groups: heatmap-based methods \cite{heatmap2,heatmap3,heatmap1}, regression-based methods \cite{reg_DEKR,reg_RLE,reg1}, and query-based methods \cite{query1,query_PETR,query2}.
For example, PETR\cite{query_PETR} introduced multiple pose queries to reason about a set of full-body poses, with a joint decoder to refine using kinematic relationships.
RLE \cite{reg_RLE} is a novel regression paradigm using Residual Log-likelihood Estimation to capture distributional changes and could facilitate the training process.
SWAHR \cite{heatmap_SAHR} adopted scale-adaptive heatmap estimation, which adjusts the standard deviation for each keypoint adaptively, making it more robust to varying scales and ambiguities.
DEKR \cite{reg_DEKR} designed a multi-branch structure for disentangled keypoint regression, which enables the model to focus on keypoint regions and improve performance.
Although existing methods have achieved great success in locating keypoints for specific classes, they cannot directly be applied for novel classes, particularly when the target classes have different numbers or types of keypoints.

\subsection{Category-Agnostic Pose Estimation}
Transfer learning is effective to learn novel classes, and thus various methods \cite{Lu1, Lu4, Yang1, Yang2, Ru2, weakshotcls, weakshotseg,jiangtongtrans,hutrans} have been proposed for extending the class scope for pose estimation models.
Specifically in few-shot learning, previous methods have primarily concentrated on specific domains, such as facial images \cite{face1, face2}, clothing images \cite{Metacloth}, or animal images \cite{AnimalZSKD1, AnimalZSKD2, AnimalZSKD3, AnimalZSKD4}.
For more diverse categories, POMNet \cite{POMNet} elaborated a large-scale dataset including 100 classes and introduced a keypoint matching framework to locate target keypoints.
CapeFormer \cite{CapeFormer} improved similarity modeling within above matching pipeline and further refined of each keypoint using a sophisticated transformer decoder.
SCAPE \cite{SCAPE} proposed global keypoint feature perceptor and keypoint attention refiner to locate target keypoints with self-attention layers.
Besides, Lu \emph{et al.} \cite{Lu1} introduced a flexible few-shot scenario that includes both novel/base classes and novel/base keypoints.
Although existing methods \cite{ESCAPE, SDPNet, DiffusionCAPE, Lu2, Lu3} have greatly advanced CAPE, how to extract FGSA features from both support and query images and how to better padding keypoints remain unexplored.
In this paper, we propose a recurrent framework to flexibly use deformable attention to mine FGSA features from support and query images, and also propose keypoint mixup padding for CAPE.

\subsection{Attention Mechanism}
Attention mechanism has been widely applied in vision tasks and achieved great success.
Existing attention mechanism could be roughly categorized into two prevalent forms: importance weight and spatial transformation. 
Importance weight is a straightforward form of attention, including spatial attention\cite{wang2017residual,jetley2018learn}, channel attention \cite{hu2018squeeze}, and the combination of both\cite{fu2019dual,woo2018cbam,jiangtongatt,huatt}. 
Spatial transformation is a special form of attention as discussed in \cite{jaderberg2015spatial,zhu2019deformable}, \emph{e.g.}, Spatial Transformer Network \cite{jaderberg2015spatial} and Deformable ConvNet (DCN) \cite{dai2017deformable}. 
Recently, multi-head attention \cite{vaswani2017attention} has shown remarkable effects in a wide range of vision tasks, but suffers from efficiency issues when applied to high-resolution feature maps.
To this end, deformable attention \cite{DETR} uses spatial transformation to effectively mine features around reference points, enabling modules to mine fine-grained features on multi-scale feature maps. 
In this paper, we design our feature miner upon deformable attention module \cite{DETR} to recurrently mine fine-grained features from support and query images, and meanwhile set reference points according to links to extract structure-aware features. 

\section{Method}
In this section, we first formally describe the task setting of class-agnostic pose estimation (CAPE) in Sec.~\ref{sec:setting}.
Afterwards, we introduce the overall pipeline of our framework in Sec.~\ref{sec:overall}.
Then, we respectively introduce the details of FGSA mining module in Sec.~\ref{sec:mining} and keypoint mixup padding and Sec.~\ref{sec:padding}.
Finally, we describe the training and inference pipeline in Sec.~\ref{sec:traintest}.

For brevity of description, we use non-bold letter to denote scalar, and employ bold letter to represent vector/matrix/tensor.
We adopt subscript to indicate the variable source and use square bracket to show the index in variable, \emph{e.g.}, $\bm{P}_s[k]$ means the $k$-th keypoint on support image.
Additionally, we employ $[\cdot;\cdot]$ to represent the concatenation of two variables.

\subsection{Task Setting} \label{sec:setting}
Category-agnostic pose estimation (CAPE) aims to locate target keypoints of query image for any novel category, where the target keypoints are determined by a few annotated support images of the same category.
Formally, $N$-shot setting means there are $N$ support images available, which could be formulated as:
\begin{equation}
    \bm{P}_q=f_{cape}(\bm{x}_q,\{\bm{x}_{s_{n}}\}_{n=1}^N,\{\bm{P}^*_{s_{n}}\}_{n=1}^N,\bm{L}^*_c),
\end{equation}
where $\bm{x}_q\in \mathbb{R}^{H\times W \times 3}$ and $\{\bm{x}_{s_n}\}_{n=1}^N$ are the query image and $N$ support images from category $c$.
The target keypoints to be estimated $\bm{P}_q$ and the $n$-th support keypoints $\bm{P}_{s_n}$ have the same size, \emph{i.e.}, $\bm{P}_q \in \mathbb{R}^{K_c\times 2}$, where $K_c$ is the keypoint number of category $c$.
Additionally, $\bm{L}^*_c \in \mathbb{R}^{K_c\times K_c}$ is a binary matrix to link keypoints to pose of category $c$, \emph{i.e.}, $\bm{L}^*_c[i,j]$ means the link existence between the $i$-th and the $j$-th keypoints.

To learn and evaluate above function $\mathcal{F}_{cape}$, all categories in benchmark are split into base categories $\mathcal{C}^b$ and novel categories $\mathcal{C}^n$, where $\mathcal{C}^b \cap \mathcal{C}^n = \emptyset $.
In the training stage, the query and support images come from only base classes, \emph{i.e.}, $c\in \mathcal{C}^b$.
In the test stage, images come from novel classes, \emph{i.e.}, $c\in \mathcal{C}^n$.
For simplicity, we first describe our framework in $1$-shot setting, \emph{i.e.}, $\bm{P}_q=f_{cape}(\bm{x}_q,\bm{x}_s,\bm{P}^*_s, \bm{L}^*_c)$, and then introduce the extension to multiple support images.

\begin{figure*}[h]
\begin{center}
\includegraphics[width=1\linewidth]{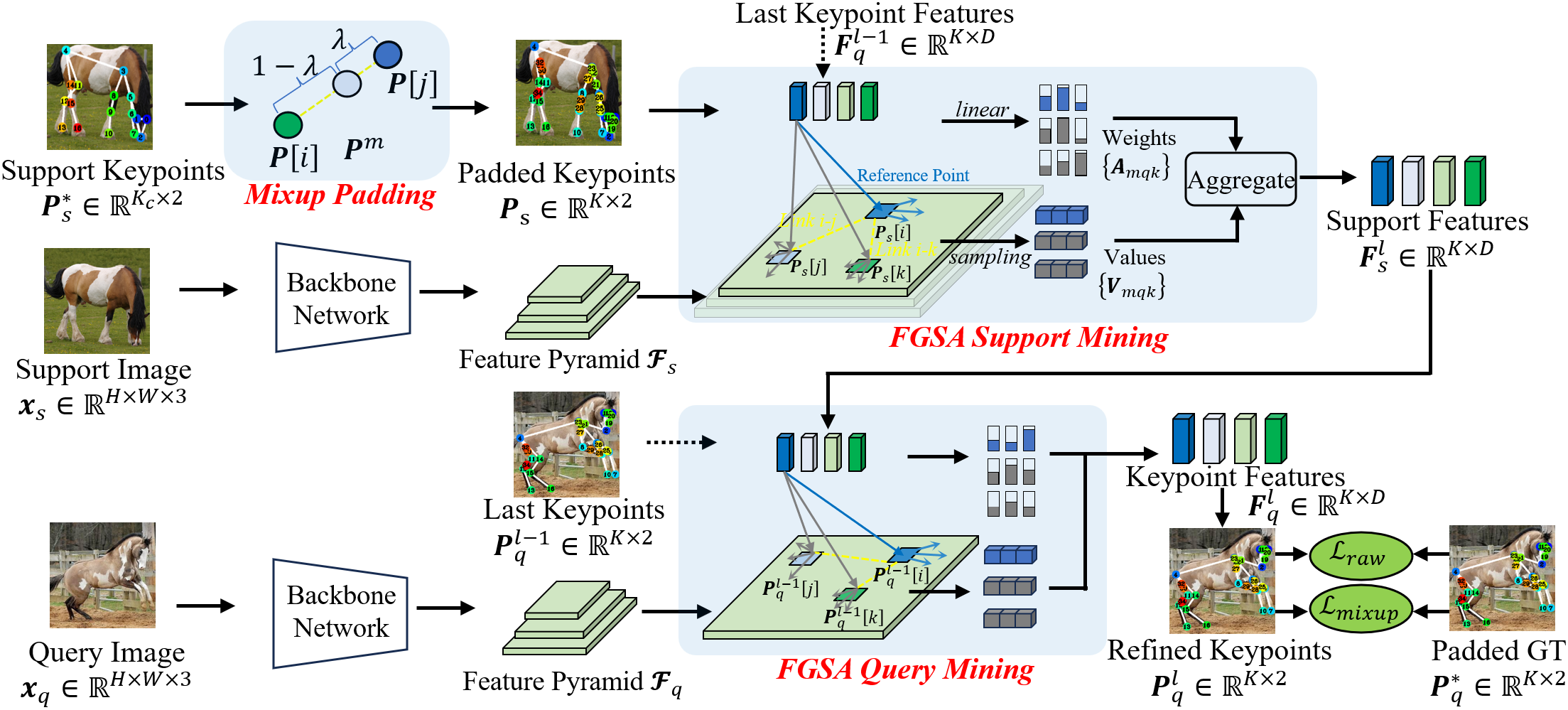}
\end{center}
\caption{
% The illustration for the detailed architecture of the $l$-th layer in our recurrent framework. 
The architecture of our $l$-th recurrent layer. 
Firstly, support keypoints $\bm{P}^*_s$ is aligned by mixup padding, which produces padded keypoints $\bm{P}_s$.
Then, $f_{miner-s}$ mines FGSA features around $\bm{P}_s$ on support image to extract support features $\bm{F}^l_s$ according to the last keypoint features $\bm{F}^{l-1}_q$.
Afterwards, $f_{miner-q}$ mines FGSA features around the last keypoints $\bm{P}^{l-1}_q$ on query image to extract keypoints $\bm{F}^l_q$, which produces target keypoints $\bm{P}^l_q$.
By recurrently updating $\bm{F}^l_s$, $\bm{F}^l_q$ and $\bm{P}^l_q$, our framework finally estimates precise keypoints $\bm{P}^L_q$.
}
\label{fig:framework}
\end{figure*}

\subsection{Overall Pipeline} \label{sec:overall}
Considering that CAPE is a pixel-level estimation task and the target keypoints are determined by both query and support images, our proposed framework recurrently mines fine-grained and structure-aware (FGSA) features from both support and query images, and thus could estimate more precise target keypoints on query images.

Specifically, our framework consists of $L$ stacked layers, and the detailed architecture of single layer is illustrated in Fig.~\ref{fig:framework}.
For the $l$-th layer, we denote the support features as $\bm{F}^{l}_s\in \mathbb{R}^{K\times D}$, which contain the semantic information of $K$ target keypoints extracted from support image and support keypoints.
We denote the keypoint features as $\bm{F}^{l}_q\in \mathbb{R}^{K\times D}$, which store the specific information of $K$ target keypoints on query image determined by support features.
The recurrent pipeline of updating support features $\bm{F}^{l}_s$ and keypoint features $\bm{F}^{l}_q$ are described as follows.
 
Given the support image $\bm{x}_s$, support keypoints $\bm{P}^*_{s}$, category-wise link $\bm{L}^*_c$ and query image $\bm{x}_q$, our framework firstly use mixup padding strategy to align the keypoint number $K_c$ to a unified number $K$, formulated as:
\begin{equation} \label{eqn:mixup}
    [\bm{P}_s;\bm{L}_c]=f_{mixup}(\bm{P}^*_{s}, \bm{L}^*_c, \alpha),
\end{equation}
where $\alpha$ is the mixup parameter as in \cite{mixup}, and $\bm{P}_s \in \mathbb{R}^{K\times 2}$ and $\bm{L}_c \in \mathbb{R}^{K\times K}$ are the padded keypoints and links.
The details about our proposed $f_{mixup}$ reminds to Sec.~\ref{sec:padding}.
Meanwhile, our framework uses a backbone network to extract multi-scale feature maps (pyramid) from support and query images, denoted as $\bm{\mathcal{F}}_s$ and $\bm{\mathcal{F}}_q$ respectively.

Afterwards, our framework produces support features $\bm{F}^{l}_s\in \mathbb{R}^{K\times D}$ by complementing previous keypoint features $\bm{F}^{l-1}_q$ with the FGSA features mined from support feature pyramid $\bm{\mathcal{F}}_s$, which could be summarized as:
\begin{equation} \label{eqn:miner_s}
    \bm{F}^{l}_s=f_{miner-s}(\bm{F}^{l-1}_q, \bm{\mathcal{F}}_s, \bm{P}_s, \bm{L}_c),
\end{equation}
where $f_{miner-s}$ denotes our proposed module mining FGSA features based on support keypoints $\bm{P}_s$ and category-wise links $\bm{L}_c$, and the details remain to Sec.~\ref{sec:mining}.
In this way, the support features $\bm{F}^{l}_s$ are also extracted according to keypoints features $\bm{F}^{l-1}_q$, and thus provide more targeted support information to refine target keypoints.

Based on the extracted support features $\bm{F}^{l}_s$, our framework uses another FGSA feature miner $f_{miner-q}$ to produce keypoint features $\bm{F}^{l}_q\in \mathbb{R}^{K\times D}$ by retrieving matched FGSA features with in query feature pyramid $\bm{\mathcal{F}}_q$, as:
\begin{equation} \label{eqn:miner_q}
    \bm{F}^{l}_q=f_{miner-q}(\bm{F}^{l}_s, \bm{\mathcal{F}}_q, \bm{P}_q^{l-1}, \bm{L}_c),
\end{equation}
where $f_{miner-q}$ mines FGSA features based on the last keypoints $\bm{P}_q^{l-1}$ and category-wise links $\bm{L}_c$.
In the $l$-th layer, the target keypoints $\bm{P}_q^{l}\in \mathbb{R}^{K\times 2}$ are estimated by:
\begin{equation}\label{eqn:delta}
    \bm{P}_q^{l}=\sigma(\sigma^{-1}(\bm{P}_q^{l-1})+f_{mlp}(\bm{F}^{l}_q)),
\end{equation}
where $\sigma$ is the Sigmoid function, $f_{mlp}$ is a light-weight MLP, and the keypoint predicting follows the incremental refinement in previous works \cite{CapeFormer,MetaPoint}.

To launch our recurrent framework, we set $\bm{F}^{0}_q$ using the heatmap pooled features of $\bm{P}_s$ and $\bm{\mathcal{F}}_s$, and set $\bm{P}^{0}_q$ using the mid-value $0.5$.
By recurrently applying Eqn.~\ref{eqn:miner_s}, Eqn.~\ref{eqn:miner_q} and Eqn.~\ref{eqn:delta}, our framework recurrently updates $\bm{F}^{l}_s$, $\bm{F}^{l}_q$ and $\bm{P}_q^{l}$, and finally obtains precise target keypoints, \emph{i.e.}, $\bm{P}_q^{L}$. 
The module details are introduced in following sub-sections.

\subsection{FGSA Feature Mining} \label{sec:mining}
To mine fine-grained and structure-aware (FGSA) features to benefit CAPE task, we design a universal module based on deformable attention \cite{DETR}, which could be applied on both support and query images.

As a flexible module, each attention head in deformable attention \cite{DETR} learns multiple sampling offets to adaptively aggregate features from multi-scale feature maps $\bm{\mathcal{F}}$ around the reference point $\bm{p} \in \mathbb{R}^{2}$.
We formulate the function of single attention head as:
\begin{equation} \label{eqn:atthead}
    \bm{f}'=f_{att}(\bm{f}, \bm{\mathcal{F}}, \bm{p}),
\end{equation}
where $\bm{f}$ and $\bm{f}' \in \mathbb{R}^{D}$ are the query and output feature vector.
The internal details (\emph{e.g.}, sampling offets and attention weights) could be found in \cite{DETR}, and thus omitted for brevity. 
Based on Eqn.~\ref{eqn:atthead}, our module naturally mines fine-grained features from feature pyramid using support keypoints or target keypoints as reference points.

The original attention module \cite{DETR} uses identical reference point for multiple attention heads, while we propose to use $\bm{L}$-adapted reference points to capture structure-aware features.
As aforementioned, our module is summarized as $f_{miner}(\bm{F},\bm{\mathcal{F}},\bm{P},\bm{L})$, where $\bm{F}\in \mathbb{R}^{K\times D}$, $\bm{P}\in \mathbb{R}^{K\times 2}$ and $\bm{L}\in \mathbb{R}^{K\times K}$ denote queries, keypoints and links.
For the $M$ attention heads of $k$-th query, the respective $M$ reference points $\mathcal{P}_{k}\in \mathbb{R}^{M\times 2}$ are derived via Breadth-First Search in the graph defined by $\bm{L}$ and starting point $\bm{P}[k]$.
And the process of our $f_{miner}$ for the $k$-th query is: 
\begin{equation}
      \bm{F}'[k]=\bm{F}[k]+\sum_m^M \bm{W}_m \cdot f_{att[m]}(\bm{F}[k], \bm{\mathcal{F}}, \mathcal{P}_{k}[m]),
\end{equation}
which can naturally extend to $K$ queries $\bm{F}\in \mathbb{R}^{K\times 2}$ as \cite{DETR}.

Therefore, our module flexibly utilizes deformable attention heads to extract fine-grained and structure-aware features, and thus can benefit the feature extraction in both support and query images, \emph{i.e.}, Eqn.~\ref{eqn:miner_s} and Eqn.~\ref{eqn:miner_q}.

\subsection{Keypoint Mixup Padding} \label{sec:padding}
In CAPE, different classes could have different numbers of keypoints, and thus keypoint padding for aligning keypoint number is inevitable.
Inspired by Mixup~\cite{mixup} and DensePose \cite{guler2018densepose}, we propose keypoint mixup padding, which could provide richer supervision and denser semantics than the zero padding used in existing works~\cite{POMNet, CapeFormer, MetaPoint}.

As aforementioned in Eqn.~\ref{eqn:mixup}, our $f_{mixup}$ pads $K_c$ keypoints to $K$ keypoints according to class-wise link $\bm{L}^*_c$. 
Firstly, we random sample $K-K_c$ keypoint pairs from all linked keypoint pairs.
Afterwards, for each sampled keypoint pair $\bm{P}^*[i]$ and $\bm{P}^*[j]$, we individually sample a $\lambda \sim {\rm Beta}(\alpha,\alpha)$ to mix them up:
\begin{equation}
    \bm{P}[k]=\lambda \cdot \bm{P}^*[i]+ (1-\lambda)\cdot \bm{P}^*[j],
\end{equation}
where $\bm{P}[k]$ denotes the padded keypoint.
% Besides, for the occluded keypoints, and their mixup keypoints are also set as occluded, \emph{i.e.}, with zero weights/visibilities.
Then, we collect all padded keypoints on the same link, and sequentially link them to replace the original link, \emph{e.g.}, $N+1$ new links for $N$ padded keypoints.
Finally, we obtain the padded keypoints $\bm{P} \in \mathbb{R}^{K\times 2}$ and links $\bm{L} \in \mathbb{R}^{K\times K}$.

\begin{figure}[h]
\begin{center}
\includegraphics[width=1\linewidth]{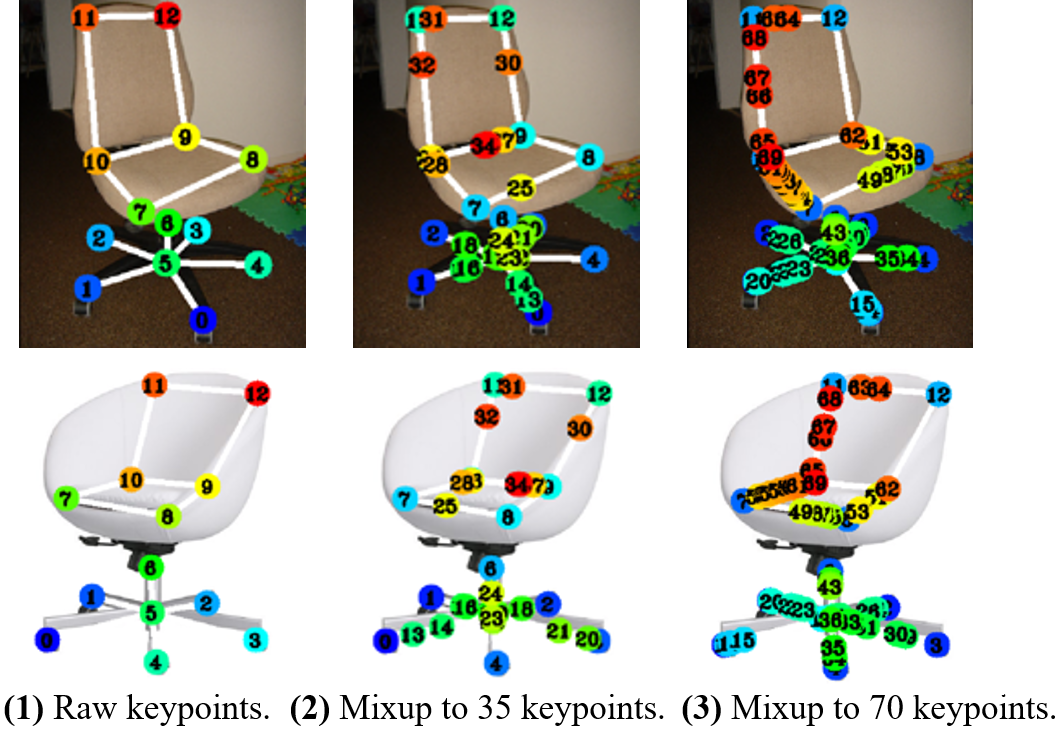}
\end{center}
\vspace{-12pt}
\caption{
Illustration of keypoint mixup padding, \emph{i.e.}, padding $16$ support keypoints (upper) and target keypoints (bottom) to $35$ or $70$ keypoints.
Therefore, our padding strategy could provide richer superivsion and enbale model to learn denser semantics.
}
\label{fig:mixup}
\end{figure}

In the training stage, the support keypoints and GT target keypoints should keep consistent, and thus we apply the same $\lambda$ in each keypoint mixup.
In the inference stage, we use uniform padding by generating equal division points on various links for a stable output.
The examples of keypoint mixup padding are illustrated in Fig.~\ref{fig:mixup} using $K=35$ or $K=70$ with $\alpha=1.0$. 
As we can see, the padded keypoints generally distrubute appropriately over the object structures, and the padded support keypoints and padded target keypoints on query image also keep the semantic consistency between support and query images.
Therefore, our model could learn denser semantics to facilitate CAPE.

\subsection{Training and Inference} \label{sec:traintest}
By our framework (Sec.~\ref{sec:overall}) and internal modules (Sec.~\ref{sec:mining} and Sec.~\ref{sec:padding}), we obtain $L$ target keypoints $\{\bm{P}^l_q\}_{l=1}^L$ for query image $\bm{x}_q$.
Our full training objective is twofold:
\begin{equation}
    \mathcal{L}_{full}=\mathcal{L}_{raw}+\beta \cdot \mathcal{L}_{mixup},
\end{equation}
where $\beta$ is a hyper-parameter for balancing.
Specifically, the first objective $\mathcal{L}_{raw}$ supervises the predictions corresponding to raw keypoints as in previous work \cite{CapeFormer}:
\begin{equation}
    \mathcal{L}_{raw}=\frac{1}{L} \sum_{l=1}^L \sum_{k=1}^{K_c} \Big| \bm{P}^l_q[k]-\bm{P}^*_q[k] \Big|_1,
\end{equation}
where $K_c$ is the raw keypoint number before padding, and $\bm{P}^*_q$ denotes the GT keypoints after padding.
Similarly, the second objective $\mathcal{L}_{mixup}$ is:
\begin{equation}
    \mathcal{L}_{mixup}=\frac{1}{L} \sum_{l=1}^L \sum_{k=K_c+1}^{K} \Big| \bm{P}^l_q[k]-\bm{P}^*_q[k] \Big|_1,
\end{equation}
which supervises on the $K-K_c$ padded keypoints.

In inference, our model outputs the result of the last layer $\bm{P}^L_q$ as the final estimation.
In $N$-shot setting, we mine FGSA features on $N$ support images in each recurrent layer and average $N$ mined features to obtain support features, \emph{i.e.}, $\bm{F}_s^l=\frac{1}{N}\sum_{n=1}^N{\bm{F}_{s,n}^l} $.

\begin{table*}[t]
\centering
\caption{The mPCK ($\%\uparrow$) performances of various methods in 1-shot setting and 5-shot setting on 5 splits of MP-100 dataset.
We summarize the PCK results with different threshold in Tab.~ \ref{tab:mPCK}. 
The best results are highlighted using boldface.}
\label{tab:SOTA}
\scalebox{0.95}{
\begin{tabular}{c|cccccc|cccccc} \toprule
\multirow{2}{*}{Method} & \multicolumn{6}{c|}{1-shot setting} & \multicolumn{6}{c}{5-shot setting} \\
 & Split1 & Split2 & Split3 & Split4 & Split5 & AVG &Split1 & Split2 & Split3 & Split4 & Split5 & AVG \\ \midrule
POMNet \cite{POMNet}&69.22&63.23&62.87&63.42&63.92&64.53&71.31&67.59&66.41&67.78&68.31&68.28\\
CapeFormer \cite{CapeFormer}&75.13&69.30&68.59&68.50&71.38&70.58&78.05&74.43&74.87&73.80&76.11&75.45\\
ESCAPE \cite{ESCAPE} &72.42&66.83&63.92&64.71&64.48&66.47&75.53&73.18&71.12&72.24&72.68&72.95\\
MetaPoint \cite{MetaPoint}&77.11&71.07&70.32&69.93&72.73 &72.23&79.22&75.51&76.20&75.92&77.65&76.90 \\
GraphCape \cite{Ru1}&73.47&68.45&67.61&67.32&68.02&68.97&77.21&74.11&73.25&73.28&73.09&74.19\\
SCAPE \cite{SCAPE}&77.53&71.21&70.41&69.51&73.12&72.36&79.35&76.26&76.29&76.12&77.89& 77.18\\
FMMP &\textbf{78.72}&\textbf{72.32}&\textbf{71.38}&\textbf{70.74}&\textbf{73.93}&\textbf{73.42}&\textbf{80.67}&\textbf{77.24}&\textbf{76.92}&\textbf{77.13}&\textbf{78.12}&\textbf{78.02} \\ \bottomrule
\end{tabular}
}
\end{table*}
\section{Experiments}
\subsection{Dataset, Metric, and Implementation Details}
Following previous CAPE studies, we utilize the MP-100 dataset \cite{POMNet} for both training and evaluation. This dataset spans $100$ classes organized into $8$ super-classes, making it the largest benchmark dataset for category-aware pose estimation (CAPE). MP-100 includes samples drawn from various category-specific pose estimation datasets, containing over $18$K images and $20$K annotations, with the number of keypoints varying from $8$ to $68$ across different classes. The $100$ classes in MP-100 are divided into non-overlapping training, validation, and test sets in a $70:10:20$ ratio.
To ensure that evaluation categories remain unseen during training, the dataset is further split into five mutually exclusive partitions, each maintaining this separation throughout the training and testing process.
By default, we use $K=70$, $M=8$, $L=3$, $\alpha=1$ and $\beta=0.5$ for our model.
We use Probability of Correct Keypoint (PCK) \cite{PCK} as the quantitative metric. Apart from the PCK of threshold $0.2$ as in \cite{POMNet,CapeFormer}, we also report the mPCK (of $[0.05,0.1,0.15,0.2]$) for a more comprehensive evaluation.

Generally, we implement our method upon the codebase of CapeFormer \cite{CapeFormer}, which is built based on PyTorch \cite{Pytorch} and MMPose \cite{mmpose}.
Specifically, we use ResNet-50 \cite{ResNet} pre-trained on ImageNet \cite{ImageNet} as our backbone, consisting with prior works \cite{CapeFormer,POMNet,MetaPoint}.
For easy process, we use the same backbone to extract multi-scale feature maps from query and support images as in \cite{CapeFormer}, and squeeze them to $256$ channels using $1\times 1$ convolutions, \emph{i.e.}, $D=256$.
Our data configuration also follows previous works \cite{POMNet,CapeFormer,MetaPoint}, \emph{i.e,}, cropping the target object according to its GT bounding box and resizing the image patch to $256\times 256$.
In the training stage, our data augmentations include random scaling and random rotation.
We use Adam \cite{Adam} optimizer to learn our model for $200$ epochs using batch size $16$, learning rate $1e^{-5}$.
Our experiment environment is builed on Ubuntu 20.04 system with 64 GB Intel 9700K CPU and 2 NVIDIA 4090 GPUs. 

\subsection{Quantitative Comparison with Prior Works}
\textbf{4.2.1 Comparable Baselines.}
We follow previous works \cite{CapeFormer,MetaPoint,SCAPE} and select state-of-the-art (SOTA) baselines for comparison. 
Considerring the utilizing of link annotations, we categorize the selected baselines into following two groups.
\textit{\textbf{(1) Point-based group.}}
POMNet \cite{POMNet} employs a keypoint matching framework to retrieve target keypoints on query images.
CapeFormer \cite{CapeFormer} further refines the matched keypoints with a novel transformer model.
MetaPoint \cite{MetaPoint} first estimates potential keypoints, and then selects and refines them to target keypoints.
ESCAPE \cite{ESCAPE} learns a prior over the features of keypoints, and then adapts them to target keypoints.
SCAPE \cite{SCAPE} focuses on learning high-quality attention  to boost the performance of CAPE.
\textit{\textbf{(2) Pose-based group.}}
GraphCape \cite{Ru1} treats the pose as a graph and uses a graph-based feed-forward network to extract geometrical features from keypoint feature vectors.
For a comprehensive comparison, we firstly copy the already reported metrics and then reproduce the missing metrics by released weights or codes.
We uniformly employ ResNet-50 \cite{ResNet} as the backbone network for fairness.

\begin{figure*}[ht]
\begin{center}
\includegraphics[width=0.95\linewidth]{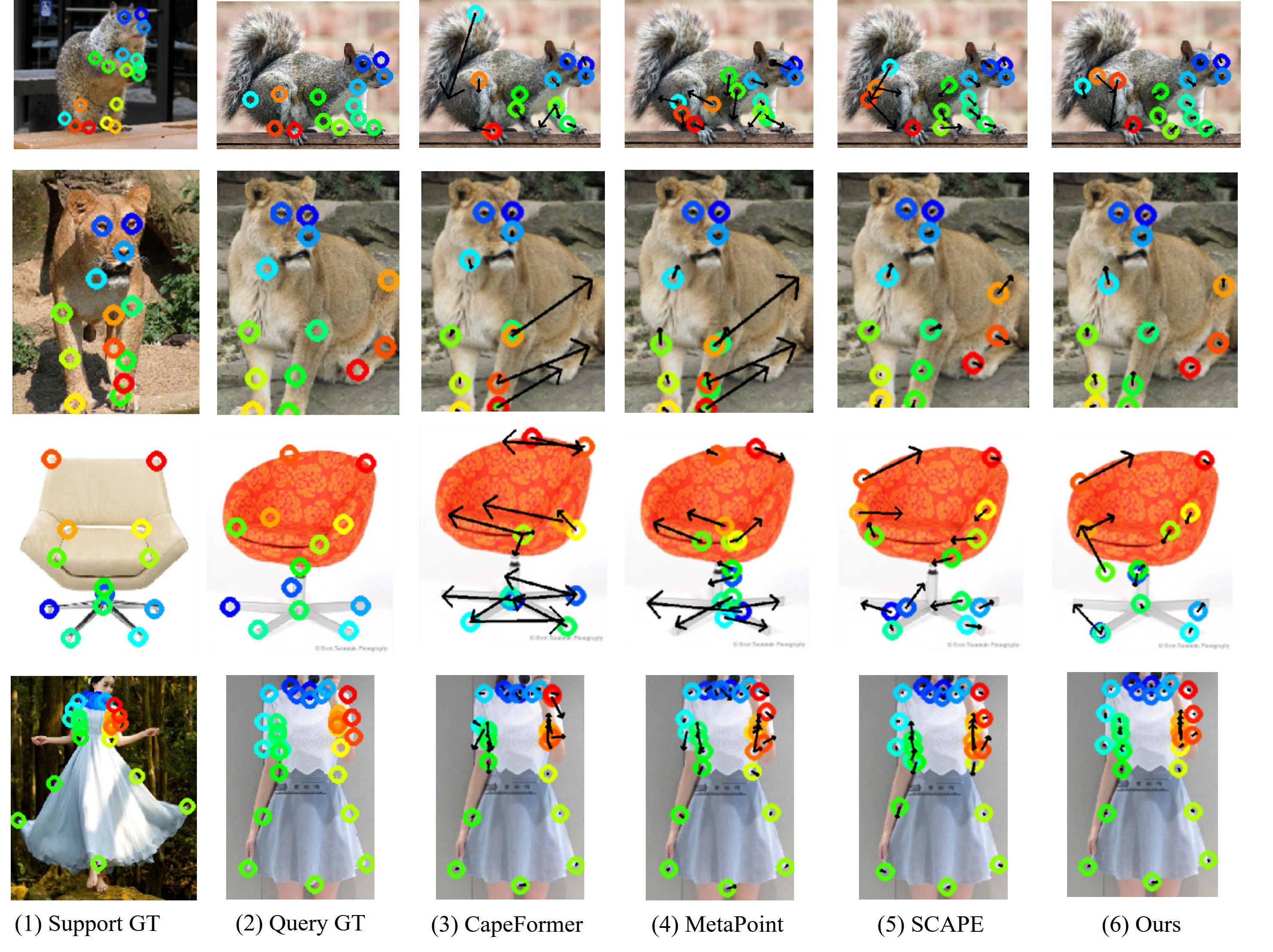}
\end{center}
\vspace{-15pt}
\caption{
Qualitative comparison.
The left two columns show the GT support and target keypoints on support and query images.
The right four columns show the target keypoints on query images estimated by various methods.
We employ graduated colors to denote the semantic classes of keypoints, and use small black arrows to indicate the deviations to GT target keypoints (\emph{i.e.}, smaller arrows are better).
}
% \vspace{-3pt}
\label{fig:viz_SOTA}
\end{figure*}

\noindent\textbf{4.2.2 Result Analysis.}
All mPCK results in 1-shot setting and 5-shot setting are summarized in Tab.~\ref{tab:SOTA}.
As a pose-based method, GraphCape \cite{Ru1} achieves favourable performances against early work \cite{POMNet} by using graph network over keypoint feature vectors, and our method further outperforms dramatically (\emph{e.g.}, +5.25 mPCK\% on Split1) by using deformable attention to mine structure-aware features from both support and query images.
Although MetaPoint \cite{MetaPoint} also employ deformable attention to mine fine-grained features from query image, our framework outperforms (\emph{e.g.}, +1.61 mPCK\% on Split1) by recurrently mining FGSA features from both support and query images.
Overall, our model achieves the optimal performances against all SOTA baselines (\emph{i.e.}, +1.06 mPCK\% in average on Split1), demonstrating the effectiveness of our method.

\noindent\textbf{4.2.3 More Detailed Comparison.}
In Tab.~\ref{tab:mPCK}, we summarize the detailed PCK results of 0.05, 0.1, 0.15 and 0.2 thresholds on Split-1 in 1-shot setting.
On the one hand, we could see that the threshold values significantly influence the performance gaps, \emph{e.g.}, SCAPE \cite{SCAPE} outperforms POMNet \cite{POMNet} $+7.4\%$ on PCK@0.2 while outperforms $+9.7\%$ on PCK@0.05, indicating that the performance measured by coarse threshold may be about saturated.
On the other hand, our method outperforms the most competitive baseline (\emph{i.e.}, SCAPE \cite{SCAPE}) by a large margin on fine threshold ($+3.2\%$ on PCK@0.05), and also achieves dramatical improvement on mPCK metric, demonstrating the effectiveness of recurrent FGSA feature mining.
\begin{table}[t]
\centering
\caption{The detailed PCK results using various thresholds (\emph{i.e.}, 0.05, 0.1, 0.15 and 0.2) and their averaged mPCK on Split1.}
\label{tab:mPCK}
\scalebox{0.85}{
\begin{tabular}{c|cccc|c} \toprule
Method & Th$0.05$ & Th$0.1$ & Th$0.15$ & Th$0.2$ & mPCK \\ \midrule
POMNet\cite{POMNet}      &44.39&68.87&79.39&84.23&69.22\\  
CapeFormer\cite{CapeFormer}  &51.03&75.17&84.87&89.45&75.13\\
ESCAPE  \cite{ESCAPE}    &48.24&72.25&82.30&86.89&72.42\\
MetaPoint\cite{MetaPoint}   &55.08&77.12&85.81&90.43&77.11\\
GraphCape \cite{Ru1}  &48.55&73.43&83.71&88.19&73.47\\
SCAPE   \cite{SCAPE}    &54.09&77.34&87.02&91.67&77.53\\
FMMP        &\textbf{57.30}&\textbf{78.48}&\textbf{87.28}&\textbf{91.82}&\textbf{78.72}\\  \bottomrule
\end{tabular}
}
\end{table}

\subsection{Qualitative Comparison with Prior Works}
In this section, we qualitatively compare our method with prior works.
Specifically, we select representative methods to visualize the estimated keypoints on the test set of dataset Split-1, including CapeFormer \cite{CapeFormer}, MetaPoint \cite{MetaPoint}, and SCAPE \cite{SCAPE}.
As shown in Fig.~\ref{fig:viz_SOTA}, our proposed method could estimate more precise keypoints for various object classes.
\emph{E.g.}, in the second row, the support keypoints are partially occluded, and our method can mine FGSA features and locate target keypoints more precisely.
Besides, our method can better locate the dense keypoints for clothing in the last row, probably due to our keypoint mixup padding strategy.
Overall, our method estimates finer and more accurate keypoints for various objects.

\begin{table}[t]
\centering
\caption{Results (mPCK) of our model in ablation study.}
\vspace{-5pt}
\label{tab:ablation}
\scalebox{0.95}{
\begin{tabular}{c|cccc|c}
\Xhline{0.8pt}
&$f_{miner-s}$ & $\mathcal{P}_{k}$ & $f_{mixup}$ & $\mathcal{L}_{mixup}$ & Split1 \\ \Xhline{0.4pt}
\#1 & - & - & - & -                                             & 69.82 \\ 
\#2 & \checkmark & - & - & -                                    & 73.18 \\
\#3 &\checkmark & \checkmark & - & -                            & 76.23\\
\#4 &\checkmark & \checkmark & \checkmark & -                   & 77.41\\
\#5 &\checkmark & \checkmark & \checkmark & \checkmark          & 78.72 \\ \Xhline{0.8pt}
\end{tabular}
}
\end{table}
\begin{table}[t]
\centering
\caption{Results (mPCK) of different versions of our model.}
\vspace{-5pt}
\label{tab:cfg}
\scalebox{0.9}{
\begin{tabular}{c|c|cccc|c} \Xhline{0.8pt}
     & Version & Split1 & Split2 & Split3 & Split4 & AVG\\ \Xhline{0.4pt}
\#1&Fledged     &78.72&72.32&71.38&70.74&73.29\\  
\#2&S$^3$Q$^3$     &77.48&71.24&70.67&70.12&72.38\\
\#3&AllLink       &76.17&70.54&70.01&69.73&71.61\\
\#4&NoneLink       &75.23&70.12&69.53&69.05&70.98\\
\#5&MixupTest       &78.35&72.08&70.94&70.43&72.95\\
\#6&ZeroTest       &76.61&70.49&70.02&69.13&71.56\\  \Xhline{0.8pt}
\end{tabular}
}
\end{table}
\subsection{Method Analysis}
\noindent\textbf{4.5.1 Ablation Study.}
To investigate the performance contributions of our modules, we gradually append modules and record results in Tab.~\ref{tab:ablation}.
Firstly, Row \#1 represent our base model, which uses heatmap pooled features $\bm{F}_q^0$ as support features and directly estimates target keypoints via $f_{miner-q}$ with conventional reference points and zero padding.
In Row \#2, we further recurrently mine features to update support features via $f_{miner-s}$, which indicates the effect of recurrent feature mining (\emph{i.e.}, +3.35\% mPCK).
In Row \#3, we set the reference points in $f_{miner-s}$ and $f_{miner-q}$ according to links to extract structure-aware features, which shows the gains of mining FGSA features from both support and query images (\emph{i.e.}, +3.05\% mPCK).
Finally, we enable keypoint mixup padding in Row \#4 and enable $\mathcal{L}_{mixup}$ in Row \#5, which could provide denser semantics and filter improper supervisions for our model.
Thus, all our modules are effective and complementary.

\noindent\textbf{4.5.2 Configurations Analysis.}
Here we analyse different versions of our model and summarize the results in Tab.~\ref{tab:cfg}.
Specifically, Row \#1 shows the standard performance of our fledged model.
Row \#2 corresponds to a conventional pipeline, \emph{i.e.}, using stacked layers to extract support features and then extract keypoint features with other stacked layers.
By comparing Row \#1 with \#2, we can see the effects of our recurrent pipeline, which could extract targeted support features according to last keypoint features.
In Row \#3 and \#4, we perturb the class-wise links to fully-connected (all-ones matrix) or only self-connected (identity matrix
). 
Correspondingly, the performances degrade due to the corrupted structure, demonstrating our model can extract structure-aware features.
In Row \#5 with Row \#6, we replace our default uniform padding in test stage to mixup padding or zero padding.
Compared with Row \#1, MixupTest slightly decreases the performance, may due to the randomness, and ZeroTest dramatically decreases due to sparser keypoints.

\begin{figure}[t]
\begin{center}
\includegraphics[width=1\linewidth]{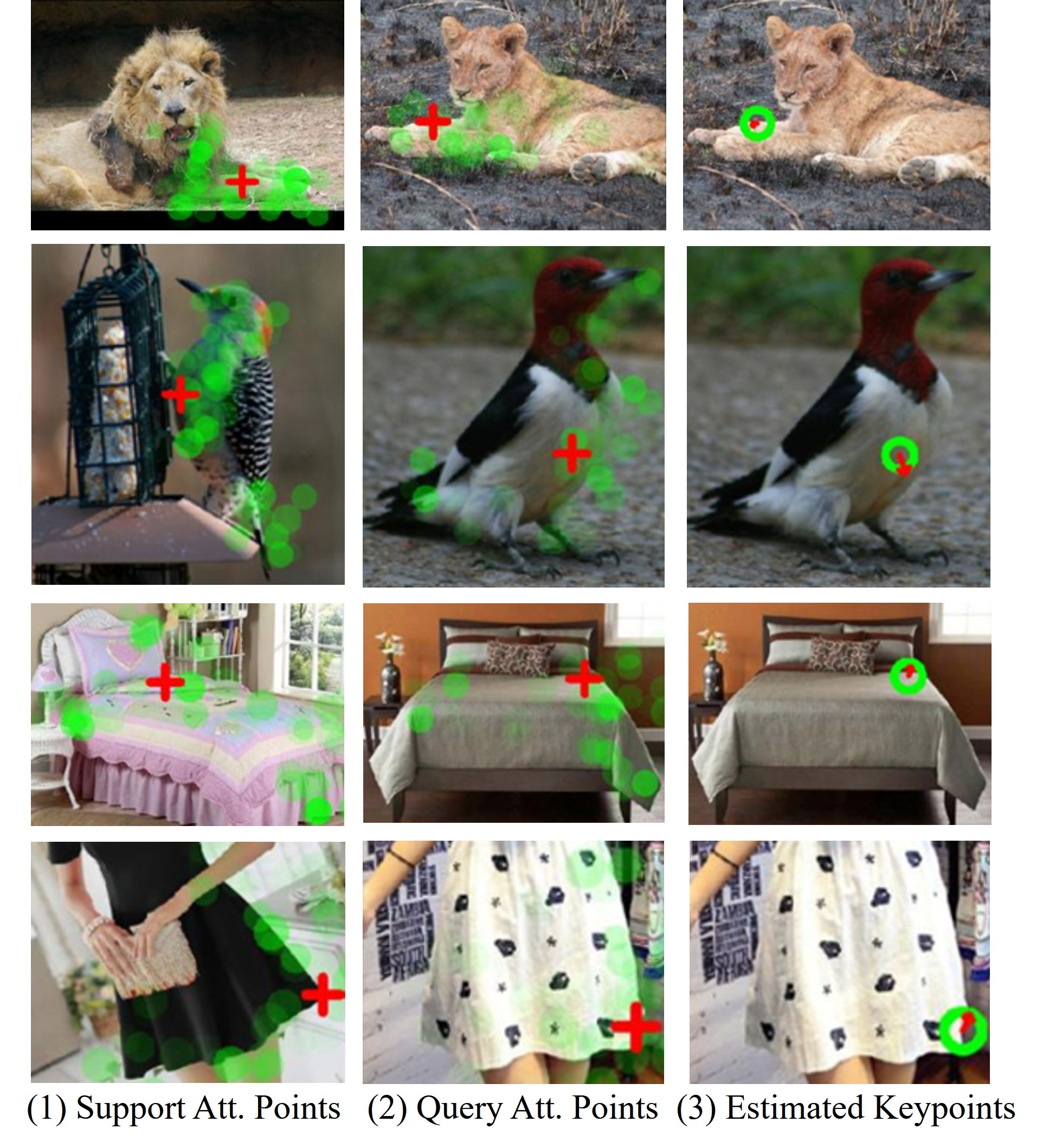}
\end{center}
% \vspace{-12pt}
\vspace{-15pt}
\caption{
Visulization for the attention on support image (col-1) and query (col-2) image, where the red cross indicates the reference point, and the green point indicates attention point with transparency as weight.
The green circle in col-3 shows estimated keypoint and the red arrow shows the deviation to GT.
}
\vspace{-5pt}
\label{fig:viz_att}
\end{figure}

\noindent\textbf{4.5.3 Qualitative Analysis.}
To intuitively understand how our model mines FGSA features, we visualize the attention points on support and query images in Fig.~ \ref{fig:viz_att}.
We can see that the attention points (\emph{i.e.}, sampling points) in deformable attention generally fit the structures of objects.
\emph{E.g.}, the attention points in the first row well fit the leg of lion in support and query images, and thus could provide structural information for support features and keypoint features.
Overall, the attention points could well capture the structures of various classes (\emph{e.g.,} lion, bird, bed and clothing), and we can find similar phenomena in other cases.

\section{Conclusion}
In this paper, we have proposed a novel framework for CAPE by recurrent feature mining and keypoint mixup padding.
Specifically, we have designed a feature mining module based on deformable attention, which could integrally mine fine-grained and structure-aware features.
Besides, we have proposed a mixup padding strategy for richer supervision and denser semantics.
We have conducted extensive experiments on MP-100 dataset, which demonstrates the effectiveness of our framework.

\section*{Acknowledgements}
This work was supported in part by the National Natural Science Foundation of China under Grants 62402201, 62271237 and U24A20220, in part by the Natural Science Foundation of Jiangxi Province of China under Grants 20242BAB26014 and 20242BAB21006, and in part by the Jiangxi Province Special Program for Cultivating Early-Career Young Scientific and Technological Talents under Grant 20244BCE52070.

{\small
\bibliographystyle{ieeenat_fullname}

}

\end{document}